\documentclass[preprint,12pt]{elsarticle}

\usepackage[utf8]{inputenc}
\usepackage[T1]{fontenc}
\usepackage{lmodern}
\usepackage{amsmath,amssymb}
\usepackage{bm}
\usepackage{xcolor}
\usepackage{hyperref}
\DeclareUnicodeCharacter{03BD}{\ensuremath{\nu}}
\DeclareUnicodeCharacter{2014}{---}
\DeclareUnicodeCharacter{2019}{'}
\DeclareUnicodeCharacter{201C}{``}
\DeclareUnicodeCharacter{201D}{''}
\DeclareUnicodeCharacter{FB02}{fl}

\journal{Acta Astronautica}

\begin{document}

\begin{frontmatter}

\title{Efficient nonlinear flame response modeling for propulsion thermoacoustic analysis using limited numerical data}

\author[addr1]{Jiawei Wu}
\author[addr1]{Teng Wang}
\author[addr1]{Jiaqi Nan}
\author[addr1]{Wang Han}
\author[addr1]{Lijun Yang}
\author[addr1,addr2,addr3]{Jingxuan Li\corref{cor1}}
\ead{jingxuanli@buaa.edu.cn}

\cortext[cor1]{Corresponding author.}

\affiliation[addr1]{organization={School of Astronautics, Beihang University},
            city={Beijing},
            postcode={100191},
            country={China}}
\affiliation[addr2]{organization={Aircraft and Propulsion Laboratory, Ningbo Institute of Technology, Beihang University},
            city={Ningbo},
            postcode={315100},
            country={China}}
\affiliation[addr3]{organization={National Key Laboratory of Aerospace Liquid Propulsion},
            city={Xi'an},
            state={Shaanxi},
            postcode={710100},
            country={China}}

\begin{abstract}

Characterizing nonlinear flame response is critical for predicting thermoacoustic instabilities in propulsion combustors, yet obtaining a comprehensive response map through high-fidelity simulations remains computationally prohibitive. This study proposes a data-driven approach for learning nonlinear flame-response dynamics from limited numerical samples. Instead of requiring exhaustive harmonic-forcing simulations, a frequency-sweeping dataset with multiple perturbation amplitudes is designed to capture the coupled effects of excitation frequency and amplitude, enabling efficient learning of the nonlinear input-output relationship between flow perturbations and heat-release-rate fluctuations. A dual-path temporal surrogate model is developed to represent nonlinear response evolution in the time domain, where complementary temporal features are extracted to retain both global response trends and local nonlinear characteristics. The proposed framework is validated using numerical simulations of a laminar premixed flame. It accurately predicts nonlinear single-frequency responses over a wide range of forcing amplitudes and frequencies, with an average mean relative error of 6.69\% for 72 independent test cases. Further evaluation using a modified $n-\tau$ model demonstrates that the framework can capture stronger nonlinear responses by increasing the diversity of the training data. This work provides an efficient alternative for constructing nonlinear flame-response models and offers a promising approach for rapid thermoacoustic stability analysis of propulsion combustors.

\end{abstract}

\begin{highlights}
\item An efficient framework models nonlinear flame response from limited data.
\item Frequency-sweeping data reduce harmonic-forcing simulation requirements.
\item A Dual-Path architecture preserves long-range and local temporal features.
\item Short-sequence sampling improves training and inference efficiency.
\item The model achieves 6.69\% average MRE over 72 single-frequency tests.
\end{highlights}


\end{frontmatter}

\section{Introduction}\label{introduction}
Combustion instability remains a major challenge in the development of high-performance propulsion systems. In liquid, hybrid, and air-breathing combustors, unsteady heat release may couple with acoustic waves, producing pressure oscillations that can reduce combustion stability, shorten component life, and restrict the operable design envelope \cite{durox2009rayleigh,poinsot2017prediction,morgans2025thermoacoustic}. This issue is particularly important for rocket engines, where injection-coupled instability in cryogenic combustors and transverse instability in multi-injector combustors continue to receive considerable attention \cite{armbruster2018experimental,ren2023experimental}. These challenges motivate the development of efficient flame-response models that are accurate enough for nonlinear thermoacoustic analysis while remaining computationally affordable for engineering applications.

A key component of low-order thermoacoustic analysis is the flame response to flow perturbations. Under small disturbances, the flame response is commonly characterized by the flame transfer function (FTF). As oscillation amplitudes increase, however, nonlinear flame dynamics become important, and the flame describing function (FDF), which depends on both forcing frequency and perturbation amplitude, is generally adopted to characterize the nonlinear response \cite{kim2016combustion, huang2009dynamics, seshadri2016reduced,kannan2016flame,zhang2021exploring}. Previous studies have also shown that delay-time effects and flame memory can significantly influence instability development \cite{bae2021effect}. Considerable progress has been made through theoretical analysis, experimental measurements, and numerical simulations \cite{fleifil1996response, lieuwen2005nonlinear, ducruix2000theoretical, noiray2008unified, krediet2013saturation, han2015simulation,liu2022effect,tian2025theoretical,liu2025analytical,gupta2025nonlinear}. Nevertheless, theoretical models are often restricted to relatively simple flame configurations, whereas experimental characterization under engine-relevant conditions remains technically challenging and resource intensive.

High-fidelity numerical simulation has become an important tool for investigating flame dynamics and supporting propulsion combustor design. However, its computational cost increases substantially when nonlinear flame response must be characterized over a broad range of forcing conditions. Conventional FDF identification typically relies on harmonic-forcing simulations conducted separately at different combinations of forcing frequency and perturbation amplitude \cite{han2015simulation,Li2017}. Such repeated simulations can become computationally demanding when extensive nonlinear response information is required. Similar challenges arise in the analysis of rocket-engine combustion instabilities, where single-pulse experiments and off-design operating conditions indicate that nonlinear high-frequency dynamics remain practically relevant \cite{umeoka2021single,kanda2023experimental}. Broadband excitation combined with system identification has substantially improved the efficiency of FTF estimation by extracting linear flame response from a single simulation \cite{polifke2014black,radack2025identification,ke2024prediction}. Extending this idea to nonlinear flame response is less straightforward because the FDF depends on both forcing frequency and perturbation amplitude \cite{liu2023models,han2015prediction,Li2015,liu2025approaches}, whereas the FTF represents the linear response and is independent of perturbation amplitude \cite{schuller2003unified,wang2009linear,li2019analytical,qiao2024flame,tian2023stretch,jaensch2018identification}. The two quantities are defined as

\begin{equation}\label{eq:FTF-FDF}
\begin{aligned}
\textrm{FTF}\left(f\right)&=\frac{q^{\prime}\left(f\right) / \bar{q}}{u^{\prime}(f) / \bar{u}},\
\textrm{FDF}\left(f,\left|u^{\prime}\right|\right)&=\frac{q^{\prime}\left(f,\left|u^{\prime}\right|\right) / \bar{q}}{u^{\prime}(f,\left|u^{\prime}\right|) / \bar{u}},
\end{aligned}
\end{equation}
where $q$ and $u$ denote the heat release rate and flow velocity at the reference location, respectively. The overbar and prime denote time-averaged and fluctuating quantities, while $f$ and $\left|u^{\prime}\right|$ represent the forcing frequency and perturbation amplitude.

Although the FDF provides a compact frequency-domain description of nonlinear flame behavior, many thermoacoustic simulations and reduced-order combustion models are naturally formulated in the time domain. Developing an efficient time-domain representation that reproduces nonlinear flame dynamics over a range of forcing conditions is therefore of practical interest. Such a representation could also facilitate future coupling with nonlinear thermoacoustic network models without requiring repeated FDF identification for each operating condition.

Recent advances in data-driven modeling provide a practical means of approximating complex nonlinear dynamical systems. In thermoacoustics, neural-network and physics-informed approaches have been explored for acoustic-field reconstruction, parameter estimation, instability prediction, limit-cycle analysis, and flame-response modeling \cite{ozan2023hard,yoko2024adjoint,wang2024early,liu2023hasr,raissi2019physics,son2024pinn,mariappan2024learning,silva2023towards,ozan2023physics,li2025unified,jatoliya2026predicting}. Several studies have also investigated the use of broadband excitation together with machine-learning techniques for nonlinear flame-response prediction \cite{jaensch2017uncertainty,tathawadekar2021modeling,yadav2022physics,wu2025extrapolation}. These studies demonstrate the potential of data-driven methods for flame-response modeling. Nevertheless, efficiently representing nonlinear flame dynamics in the time domain from limited high-fidelity simulation data remains an open challenge, particularly when prediction accuracy and computational efficiency must both be maintained.

Motivated by this need, the present work develops a data-driven framework for representing nonlinear flame response in the time domain using a limited set of numerical simulations. Rather than replacing existing flame-response characterization methods, the proposed framework is intended to provide an efficient surrogate model that complements conventional numerical simulations and can be incorporated into future thermoacoustic analyses. The present study considers a laminar premixed flame as a benchmark configuration, allowing the response-modeling problem to be investigated without additional uncertainties introduced by turbulence, spray dynamics, or complex combustor geometries.

The proposed methodology has three main features. First, frequency-sweeping excitation with multiple perturbation amplitudes is adopted for training, reducing the number of harmonic-forcing simulations required to characterize nonlinear flame response. Second, a dual-path temporal representation is employed to capture both long-term response evolution and local temporal characteristics, providing an effective representation of nonlinear flame dynamics. Third, a short-sequence sampling strategy improves computational efficiency while preserving the dominant temporal information required for response prediction. The resulting model is evaluated against independent single-frequency numerical simulations over a broad range of forcing frequencies and amplitudes.

\section{Numerical setup for data generation}
\label{sec:numerical-setting}

\begin{figure}[htbp]
\centering
\includegraphics[width=0.55\textwidth]{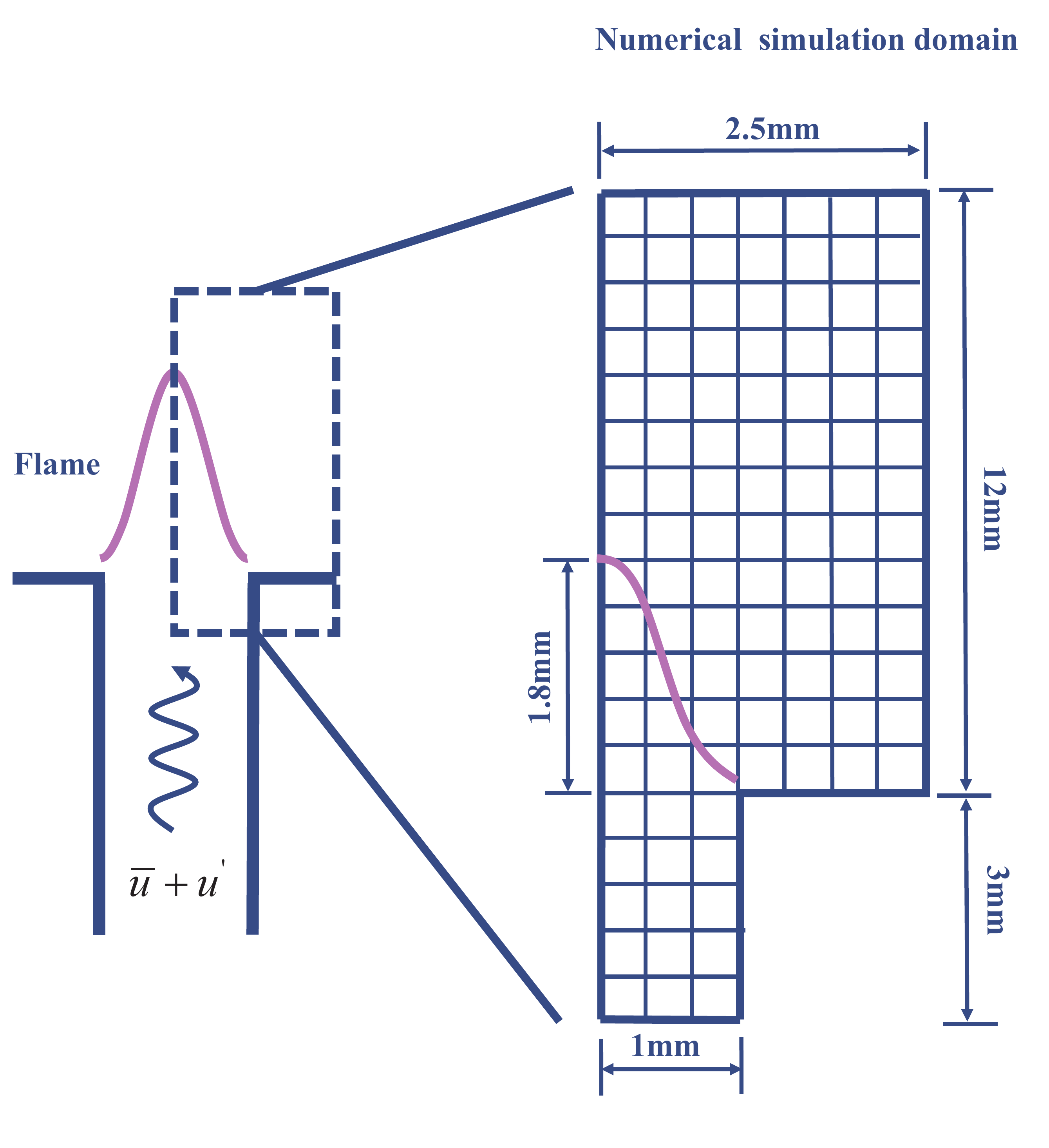}
\caption{Numerical simulation configuration.}
\label{numerical-simulation-configuration}
\end{figure}

A controlled laminar premixed flame is used to generate the nonlinear response data required for method validation. This configuration is not intended to reproduce the full complexity of an operational propulsion combustor; rather, it provides a well-defined flame-response problem in which the input velocity perturbation and output heat-release-rate perturbation can be sampled with high temporal resolution. The two-dimensional numerical simulation is performed using the open-source software OpenFOAM with the reactingFoam solver \cite{gonzalez2017numerical, yang2019reactingfoam, wang2025openfoam}. The flow is assumed to be axisymmetric and laminar. The flame operates at atmospheric conditions, and radiant heat exchange between the wall and gas is ignored. The numerical domain is shown in Fig.~\ref{numerical-simulation-configuration}. The radius of the injector outlet is 1 mm, and the injector channel length is three times the injector radius. To allow the flame to develop while reducing outlet effects, the combustion chamber length is set to 12 times the injector radius. The injector channel and injector panel, where the flame is stabilized, are maintained at 293 K. No-slip boundary conditions are used at the walls. A structured grid with 25,000 cells is used, with local refinement near the wall to resolve boundary-layer effects. The grid is nonuniform, with a cell size of 0.01 mm near the wall and a growth rate of 1.1. The simulation uses a fixed time step of $\triangle t = 1 \times 10^{-6}$ s. Methane and air are premixed at an equivalence ratio of $\phi = 0.8$, and the inlet mean velocity is $\bar{u} = 1$ m/s. Chemical kinetics are represented by a multi-step mechanism with twenty species and seventy-nine reactions \cite{petrova2006small}.
	
\section{Training dataset construction}
\label{sec:training-dataset-construction}
\begin{figure}[htbp]
\centering
\includegraphics[width=\textwidth]{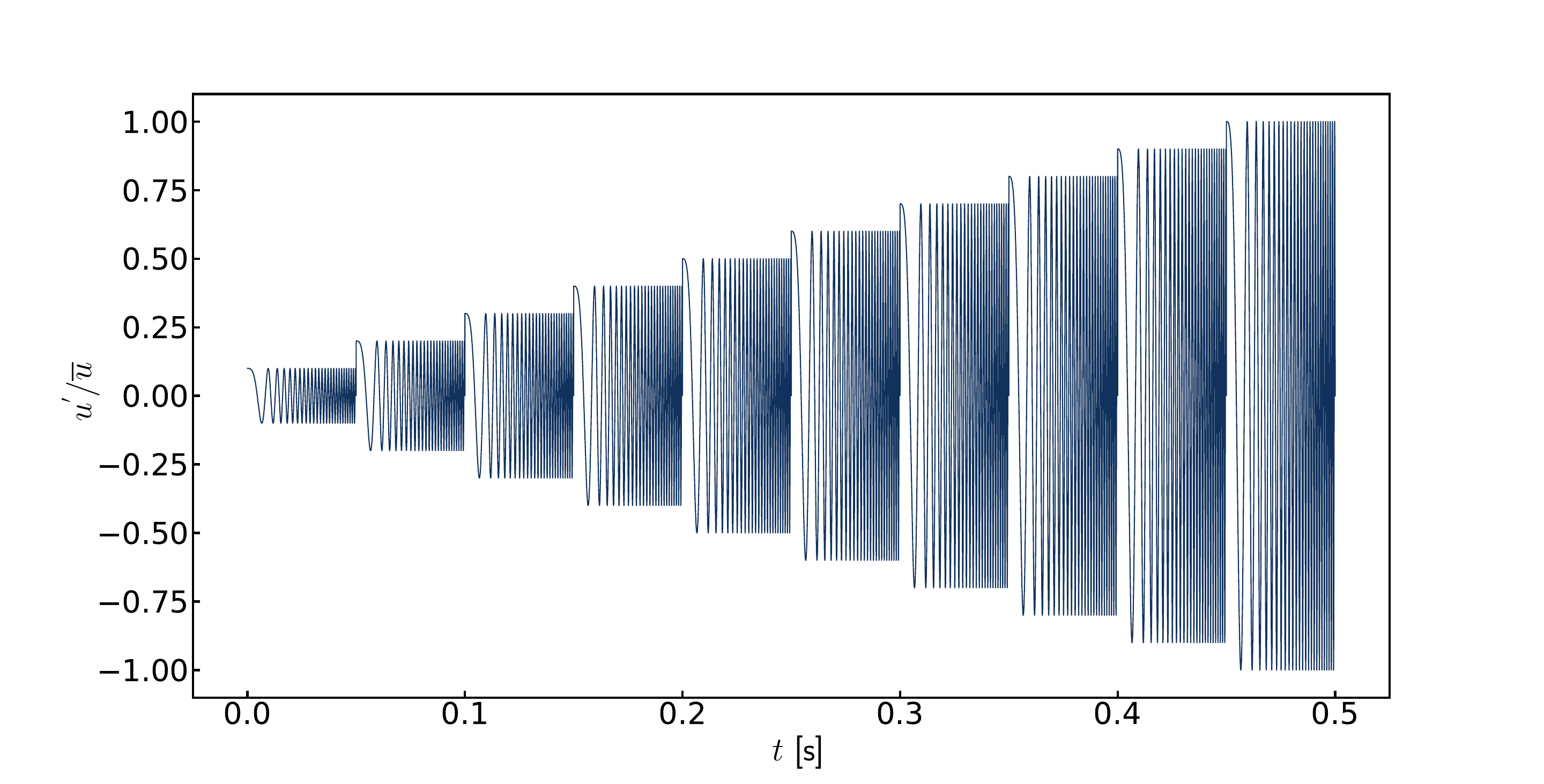}
\caption{Schematic of frequency-sweeping signals with different amplitudes.}
\label{frequency-sweeping}
\end{figure}

Efficient nonlinear response construction requires training data that contain both frequency-dependent and amplitude-dependent flame dynamics. Generating separate harmonic-forcing simulations for every frequency and amplitude pair would reproduce the cost bottleneck addressed by this study. The training dataset is therefore built from linear frequency-sweeping signals with different amplitudes, as shown in Fig.~\ref{frequency-sweeping}.
 
The normalized linearly frequency-swept velocity perturbation can be expressed as:
\begin{equation}
\begin{aligned}
\frac{u^\prime(t)}{\bar{u}} &= A \cos\left[2\pi\int_{t_1}^{t} f(t) \mathrm{d}t\right] \\
&= A\cos\left[2\pi\left(\frac{a}{2}\left(t^{2}-t_1^{2}\right)+b(t-t_1)\right) \right],
\quad t\in [t_1, t_2],
\end{aligned}
\label{Eq:frequency-sweeping}
\end{equation}
where $t_{1}$ and $t_{2}$ denote the start and end times of each linearly frequency-swept signal, while $f_{1}$ and $f_2$ denote the start and end frequencies, respectively. $\bar{u}$ and $u^\prime(t)$ denote the mean flow velocity and velocity perturbation at the burner inlet. $A$ denotes the normalized perturbation amplitude.
$a = (f_2-f_{1})/(t_{2}-t_{1})$ and $b =(t_{2}f_{1}-t_{1}f_{2})/(t_{2}-t_{1})$. The frequency therefore increases linearly with time and is given by $f(t) = a~ t + b$. 
	
\par

To cover the relevant inlet velocity perturbation amplitudes, we consider ten groups of linearly frequency-swept velocity perturbations with $A$ ranging from 0.1 to 1 in increments of 0.1. In each group, the frequency increases linearly from 10 to 1000~Hz, covering the frequency range of interest. This dataset allows the model to learn nonlinear flame dynamics across both frequency and amplitude using limited data. It also avoids the need to generate separate single-frequency signals for each amplitude--frequency combination with high-fidelity numerical simulations. Frequency-sweeping signals with different amplitudes therefore reduce the cost of dataset generation while retaining rich response information.

Each numerical simulation used in this study lasts 0.056~s and is sampled at $\Delta t=1\times10^{-6}$~s. After each simulation, we generate training sample pairs $(U^{\prime} \in \mathbb{R}^{6000},q^{\prime} \in \mathbb{R}^{1})$ following Eq.~\eqref{Eq:q-u}. With the sample length $n$ set to 6000, 50,000 training sample pairs can be generated from each group. The choice of $n$ is discussed in Sec.~\ref{short-sequence-sampling}. With ten groups, the complete training dataset contains 500,000 sample pairs.

\begin{equation}
q^{\prime}(t)=f(u^{\prime}(t-(n-1)dt), u^{\prime}(t-(n-2)dt), \ldots, u^{\prime}(t))
\label{Eq:q-u}
\end{equation}

The training and test sets are deliberately separated by signal type. The model is trained on frequency-sweeping signals with multiple amplitudes, which expose the network to a broad response manifold using limited simulation time. It is then tested on single-frequency signals at prescribed amplitudes because thermoacoustic limit cycles in propulsion and power-generation combustors often contain dominant tonal components. The predicted heat-release-rate signal is compared directly with the corresponding numerical result.

\section{Surrogate architecture for flame response modeling}
The response surrogate must reflect the temporal structure of the flame-response problem. The input is a velocity-perturbation sequence, and the target is the global heat-release-rate perturbation at a given time. Three features of this mapping are important. First, the input is a time series, so the ordering and temporal relationships among velocity samples must be retained. Sequential architectures such as Long Short-Term Memory (LSTM) networks \cite{hochreiter1997long} and Transformer networks \cite{vaswani2017attention} are therefore natural candidates for capturing this dependence.

Second, the flame response has a finite but long memory. In this study, one input velocity-perturbation sample contains 6000 points, which makes direct sequence learning inefficient \cite{bengio1994learning, pascanu2013difficulty}. Such long sequences increase the risk of vanishing gradients in LSTM models, and specialized recurrent architectures have been proposed to extend the learnable sequence length \cite{li2018independently}. Long inputs also increase the computational cost of Transformer self-attention \cite{vaswani2017attention, tay2022efficient}. A reduction of sequence length is therefore required before sequential feature extraction.

Third, low-level temporal features are important for both accuracy and generalization. Changing one velocity sample, or changing the order of two samples, can alter the heat-release-rate response. Deep networks often compress detailed input information as features propagate to deeper layers; shallow features can therefore retain local variations that are useful for resolving nonlinear flame dynamics. The proposed architecture explicitly preserves and reinjects such temporal-detail information.
	
\begin{figure}[htbp]
\centering
\includegraphics[width=\textwidth]{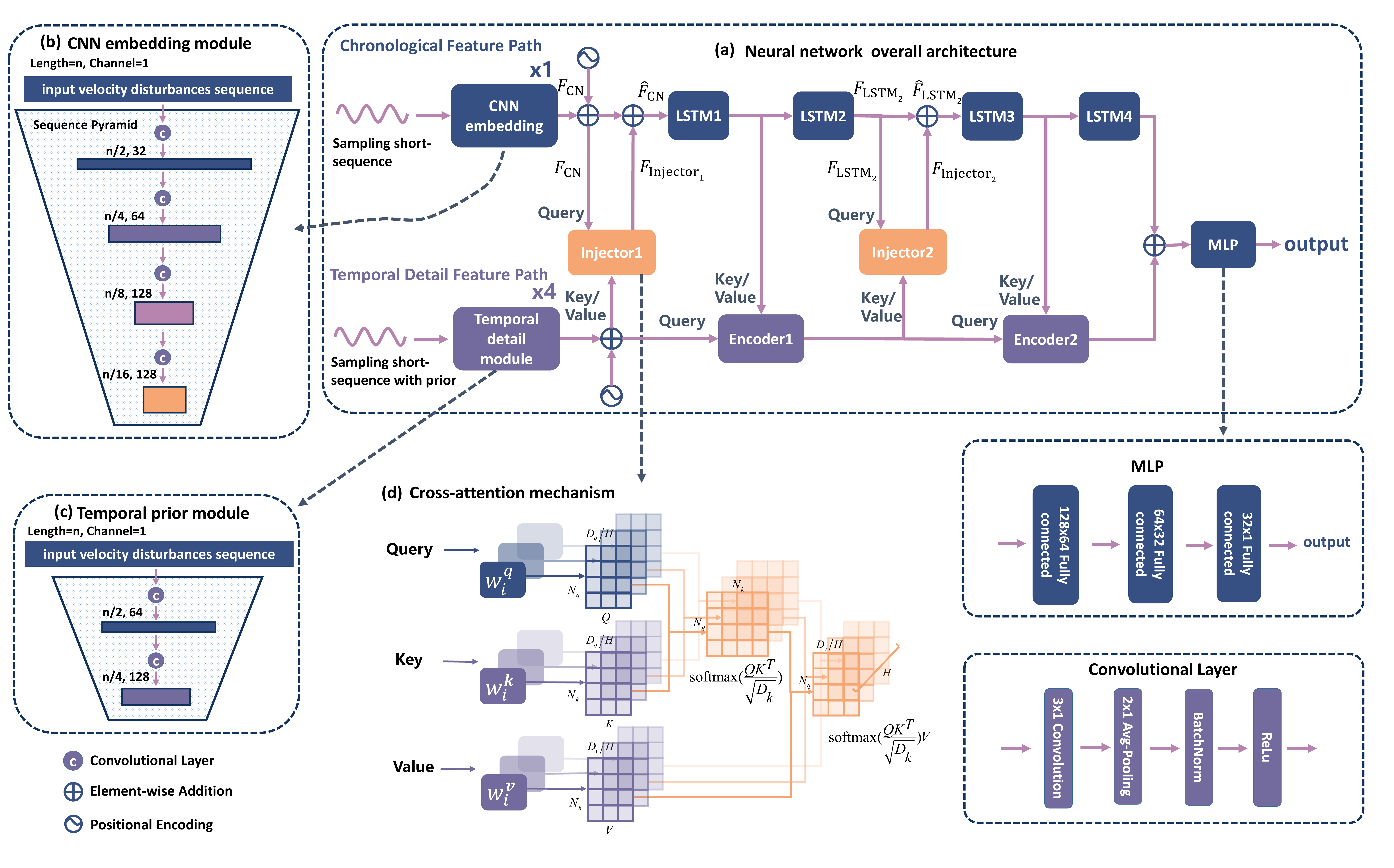}
\caption{Overall architecture of the neural network. ``Avg-Pooling'' denotes average pooling.}
\label{nerual-network-overall-architecture}
\end{figure}

Our previous work combined a Decreasing Sequence Increasing Dimension (DSID) module with a sequence module to address sequence length and temporal dependence \cite{wu2023reconstruction}. Building on that basis, the present study introduces a Dual-Path model that uses low-level temporal features to improve nonlinear response construction and generalization. Fig.~\ref{nerual-network-overall-architecture} shows the architecture. The Chronological Features Path (CFP) extracts ordered sequence features using a Convolutional Neural Network (CNN) embedding module and four LSTM layers. The Temporal Detail Feature Path (TDFP) extracts local temporal details through a temporal detail module and two encoder modules. Injector modules exchange information between the two paths. The model uses CNN \cite{lecun1998gradient}, Recurrent Neural Network (RNN) \cite{elman1990finding}, and Transformer \cite{vaswani2017attention} concepts; their basic descriptions are provided in Supplementary Material A and in the cited references \cite{lecun1998gradient,elman1990finding,vaswani2017attention}.
 
The forward propagation of sequence features in the Dual-Path model is illustrated in Fig.~\ref{nerual-network-overall-architecture}(a). The input velocity perturbation sequence contains $n$ samples, each with a feature dimension of 1. The original velocity perturbation sequence is first converted into a short sequence, as described in Sec.~\ref{short-sequence-sampling}. The resulting short sequences are then fed into two paths for learning: the CFP and the TDFP. In the CFP, the sequence first passes through the CNN-embedding module to reduce its length. The positional encoding, which adds temporal information to features \cite{vaswani2017attention}, and the features output by the Injector module are then incorporated into the sequence features from the CNN-embedding module. These sequence features are subsequently fed into four LSTM layers to extract ordered temporal features. The design motivation and details of the CFP are provided in Supplementary Material B.
		
For the TDFP, the short sequence with prior information, defined in Sec.~\ref{short-sequence-sampling}, is processed by a temporal-detail module. Positional encoding is then added to the output features from this module. These features, enriched with detailed temporal characteristics and prior information, serve as key and value inputs to the Injector module and as query inputs to the Encoder module. This study uses two Injector modules and two Encoder modules. The Injector module handles feature interaction by accepting queries from the CFP and keys and values from the TDFP, then adding the processed features back to the CFP. The Encoder module also performs feature interaction, but in the opposite direction: it accepts keys and values from the CFP and queries from the TDFP. The positions of the two Injector modules and two Encoder modules within the overall neural network are shown in Fig.~\ref{nerual-network-overall-architecture}(a). Finally, the features extracted by Encoder2 and the feature corresponding to the last step of LSTM4 are combined and fed into a Multi-Layer Perceptron (MLP) module \cite{hornik1991approximation} to produce the output. Because the desired output is the global heat release rate at a given time, the MLP module projects the learned features to a single value. The design motivation and details of the TDFP are provided in Supplementary Material C.
 
\section{Short sequence sampling}
\label{short-sequence-sampling}

The relationship between the velocity perturbation sequence and the global heat release rate is many-to-one. Specifically, all $n$ velocity perturbation samples can influence the global heat release rate at time $t$, as depicted in Eq.~\eqref{Eq:q-u}.

\begin{figure}[htbp]
	\centering
	\includegraphics[width=\textwidth]{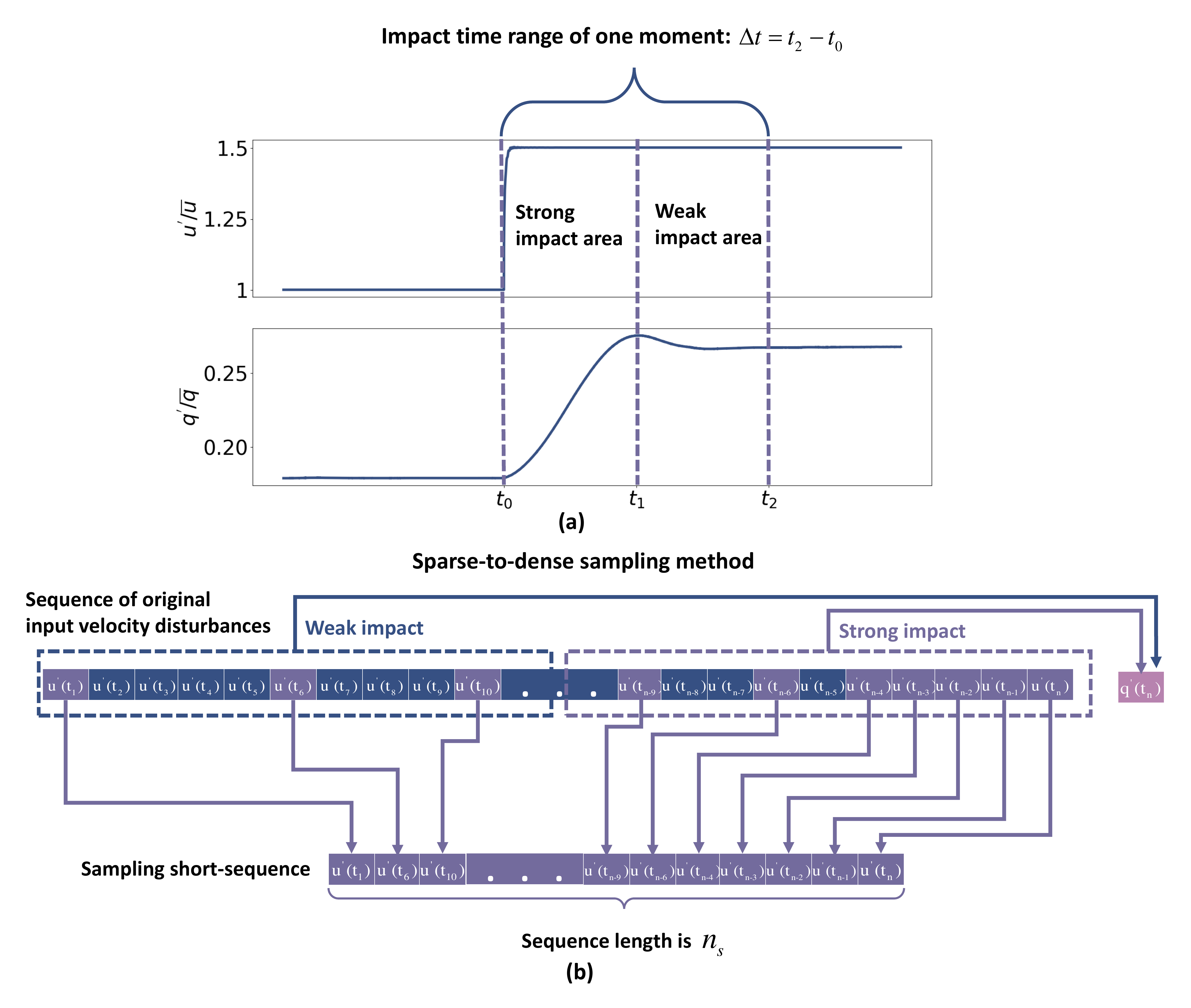}
\caption{(a) Flame dynamic response to the step signal. (b) Schematic diagram of the short-sequence sampling method; $n$ and $n_{s}$ are the lengths of the original sequence and sampled short sequence, respectively.}
	\label{Impact-time-and-sampling-method}
\end{figure}

The value of $n$ depends on the time-delay characteristics of the flame model. To determine an appropriate value of \( n \), a step signal is applied, as illustrated in Fig.~\ref{Impact-time-and-sampling-method}(a). The fluctuation interval of the heat-release-rate signal is then observed. The impact time, \( \Delta t \), is estimated as \( t_{2} - t_{0} \), and \( n \) is calculated as \( n = \frac{\Delta t}{dt} \).

The original input sequence is long because the neural network must learn the flame memory over $n$ velocity-perturbation samples. Adjacent samples change only slightly, so the sequence contains redundancy. We therefore introduce a short-sequence sampling method to reduce training and inference cost while retaining the response-relevant history. Equal-interval sampling is used for the CFP, where consecutive sampling points have the same spacing. Sparse-to-dense sampling is used for the TDFP, as illustrated in Fig.~\ref{Impact-time-and-sampling-method}(b), where the sampling interval decreases along the input sequence. Eq.~\eqref{Eq:sampling} defines the sampling indices.

Using different sampling methods in the two paths provides richer sequence information than applying one sampling strategy to both paths. Equal-interval sampling preserves the overall sequence range, whereas sparse-to-dense sampling emphasizes the later velocity perturbations that have a stronger influence on the output. Using only one of these strategies reduces the information available for learning.

\begin{equation}
\left\{\begin{matrix}
\begin{aligned}
&dn=2n/(1+n_{s})n_{s}\\
&\Delta n_i=\Delta n_{i-1}-dn\\
&index_i=\text{int}(index_{i-1}+\Delta n_i)\\
&\Delta n_1=n_{s} \times dn,\; index_1=0
\label{Eq:sampling}
\end{aligned}
\end{matrix}\right.
\end{equation}
	
In the equation above, \( n \) denotes the original sequence length, \( index_i \) denotes the sampling index of the \( i \)-th value in the short sequence, \( \text{int()} \) denotes integer truncation, and \( n_s \) denotes the length of the sampled short sequence. The sampling step \( \Delta n \) starts large and gradually decreases over time, resulting in a sparse-to-dense distribution of sampling indices. This sampling strategy is motivated by the unequal influence of the $n$ velocity perturbation samples on the global heat release rate at time \( t_n \), as illustrated in Fig.~\ref{Impact-time-and-sampling-method}(b). A velocity perturbation closer to \( t_n \) has a stronger influence on the heat release rate at that time; a more distant perturbation has a weaker influence.

This phenomenon is demonstrated in Fig.~\ref{Impact-time-and-sampling-method}(a), where a velocity step signal changes from a steady velocity of 1 to 1.5 and produces a significant heat-release-rate response. The heat release rate changes sharply from \( t_0 \) to \( t_1 \). The variation then becomes smaller, and the signal reaches a steady state after \( t_2 \). The influence time window of each velocity perturbation sample can therefore be defined as \( \Delta t = t_2 - t_0 \), with its influence on the output decreasing over this interval.

The TDFP therefore adopts sparse-to-dense sampling, using fewer points at earlier times and more points near the output time. This design incorporates \textit{a priori} knowledge of the flame response: later velocity perturbations have a stronger effect on the heat-release-rate perturbation. To validate this choice, we test single-frequency signals with amplitudes from 0.25 to 0.95 in increments of 0.1 and frequencies from 100 to 900~Hz in increments of 100~Hz, giving 72 signals in total. Compared with equal-interval sampling in the TDFP, sparse-to-dense sampling reduces the Mean Relative Error (MRE) by 1.2\%. The MRE is defined in Eq.~\eqref{Eq: MRE}, where \( q_i' \) is the \( i \)-th numerical-simulation result and \( \hat{q}_i' \) is the corresponding neural-network prediction. In this study, the original sequence length of 6000 is reduced to a short sequence of 1000. Training and inference speeds increase by factors of 15.4 and 7.7, respectively.

\begin{equation}
\text { MRE }=\frac{ \frac{1}{m} \sum_{i=1}^{m}\left | q_{i}^{\prime}-\hat{q}_{i}^{\prime} \right |}{\frac{1}{m} \sum_{i=1}^{m}|q_{i}^{\prime}|}
\label{Eq: MRE}
\end{equation}

\section{Results}
\subsection{Implementation details}
The implementation is conducted in PyTorch \cite{paszke2019pytorch}. The datasets are obtained from the numerical simulations described in Sec.~\ref{sec:numerical-setting}, and the training dataset uses the frequency-sweeping signals proposed in Sec.~\ref{sec:training-dataset-construction}. During training, BatchNorm \cite{ioffe2015batch} is added to the convolutional neural network and fully connected layers to estimate the mean and variance of each layer and adjust the data distribution. This operation stabilizes training and accelerates convergence. We use the Adam optimizer \cite{kingma2015adam} with $\beta_{1}=0.9$, $\beta_{2}=0.999$, $\varepsilon=10^{-9}$, and a weight decay of 0.01. The initial learning rate is $10^{-4}$ and is adjusted using PyTorch's MultiStepLR method, with milestones set to [40,80,90] and gamma set to 0.1. The neural-network configurations are listed in Supplementary Material Table 1. The loss function uses MSE, as shown in Eq.~\eqref{Eq: MSE}; $\hat{q}_{i}^{\prime}$ is the neural-network output, and $q_{i}^{\prime}$ is the numerical-simulation result.
\begin{equation}
\text { MSE loss }=\frac{1}{n} \sum_{i=1}^{n}\left(q_{i}^{\prime}-\hat{q}_{i}^{\prime}\right)^{2}
\label{Eq: MSE}
\end{equation}

\subsection{Nonlinear flame response construction}\label{sec: construction capability of FNR}

This section evaluates whether the Dual-Path model can reconstruct nonlinear flame responses from numerical data. The comparison uses an MLP baseline and an LSTM baseline, representing a standard feed-forward model and a standard sequence model. To avoid attributing the improvement merely to a larger parameter count, the learnable parameters of the MLP, LSTM, and Dual-Path models are kept at the same order of magnitude. The parameter counts are 1,050,379, 1,017,692, and 1,043,137, respectively, as obtained from PyTorch. The MLP has four layers with input and output feature dimensions of [6000,170], [170,128], [128,64], and [64,1], and uses a dropout rate of 0.5 as in Ref.~\cite{tathawadekar2021modeling}. The LSTM also has four layers, with feature dimensions [1,300], [300,256], [256,64], and [64,1], and uses a dropout rate of 0.3 as in Ref.~\cite{yadav2022physics}.

\begin{figure}[htbp]
\centering
\includegraphics[width=0.9\textwidth]{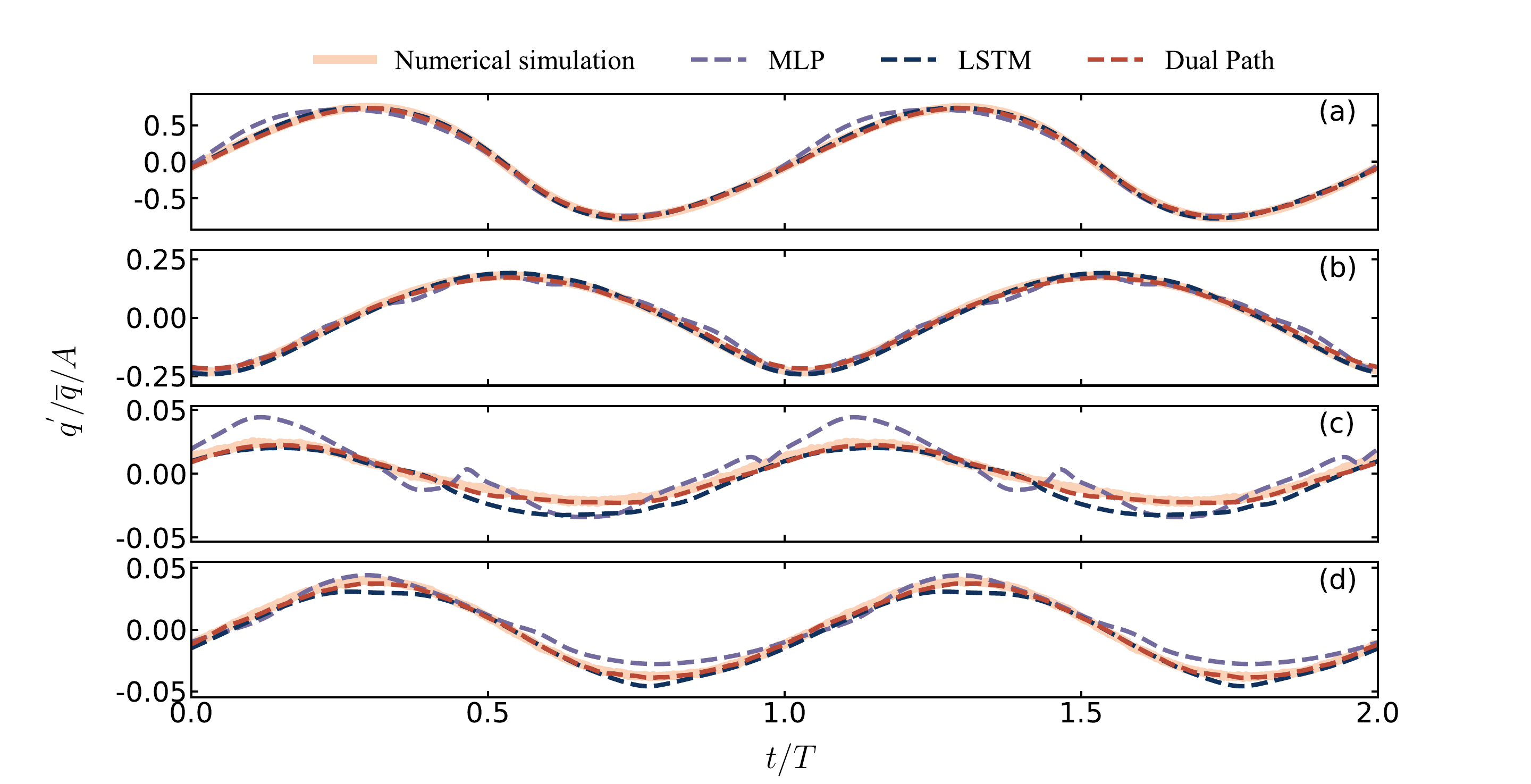}
\caption{Comparison of predictions from different neural network models when $A=0.25$. From (a) to (d), the perturbation frequencies are 200~Hz, 400~Hz, 600~Hz, and 800~Hz, respectively. The original input sequence length $n$ is 6000, and the short-sequence sampling length $n_{s}$ is set to 1000. $T$ and $A$ denote the period and amplitude of each signal, respectively.}
\label{result-0.25amplitude}
\end{figure}

\begin{figure}[htbp]
\centering
\includegraphics[width=0.9\textwidth]{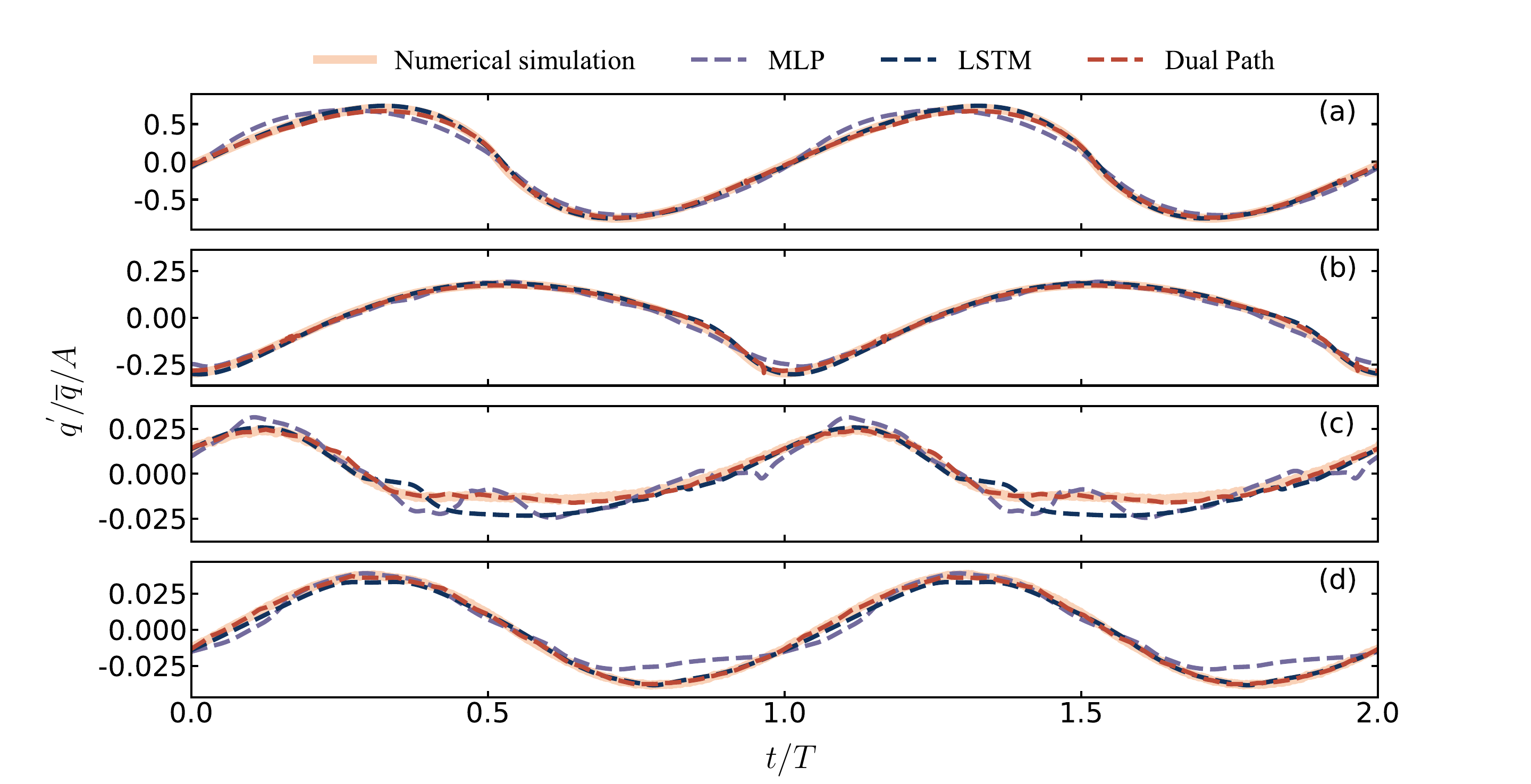}
\caption{Comparison of predictions from different neural network models when $A=0.55$. The remaining details are identical to those described in the caption of Fig.~\ref{result-0.25amplitude}.}
\label{result-0.55amplitude}
\end{figure}

\begin{figure}[htbp]
\centering
\includegraphics[width=0.9\textwidth]{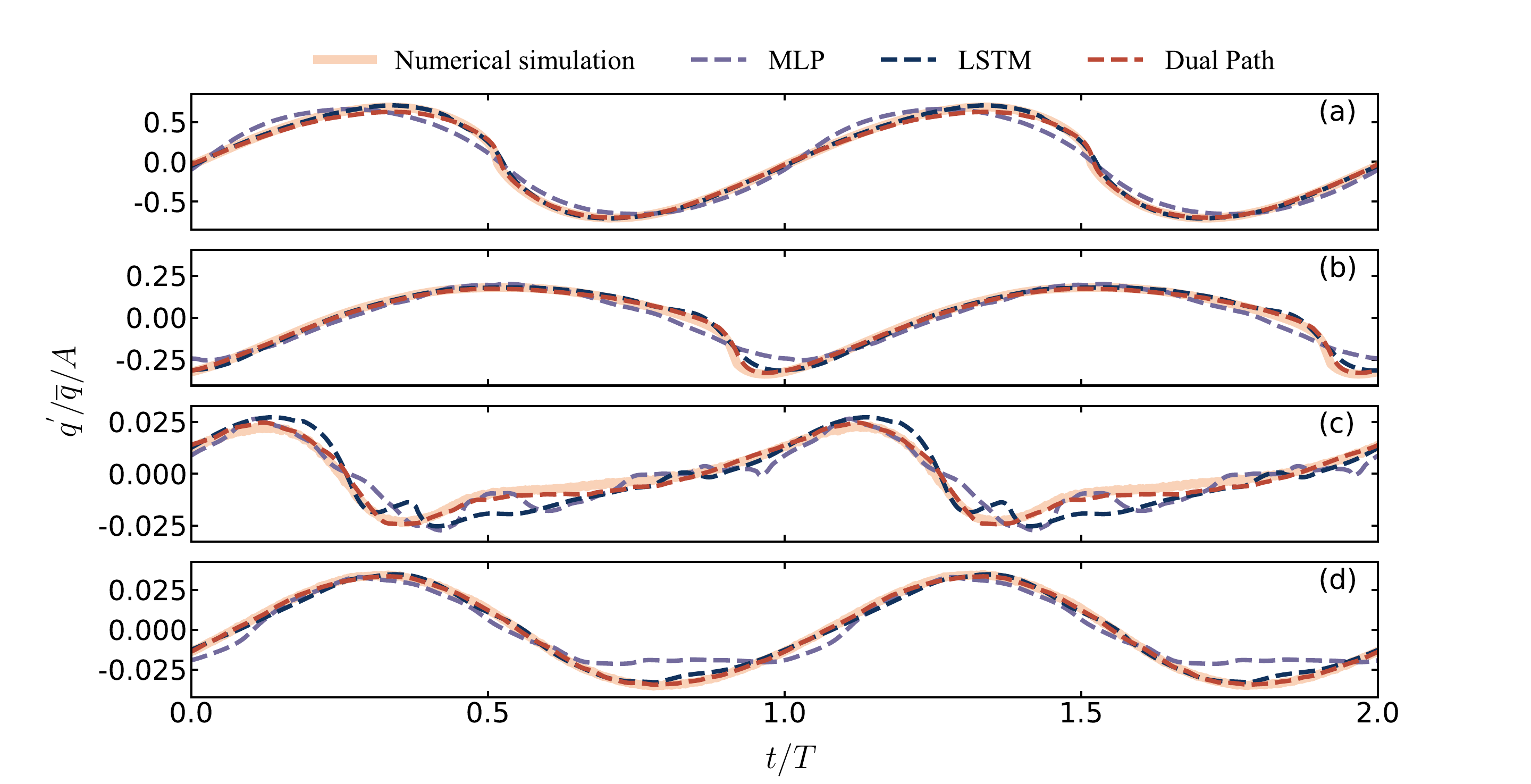}
\caption{Comparison of predictions from different neural network models when $A=0.85$. The remaining details are identical to those described in the caption of Fig.~\ref{result-0.25amplitude}.}
\label{result-0.85amplitude}
\end{figure}

Figs.~\ref{result-0.25amplitude}, \ref{result-0.55amplitude}, and \ref{result-0.85amplitude} show time-domain predictions for velocity-perturbation amplitudes of 0.25, 0.55, and 0.85 at frequencies of 200, 400, 600, and 800~Hz. The Dual-Path model tracks the nonlinear response across the tested single-frequency signals. The improvement is especially clear at 600~Hz, where the MLP and LSTM baselines show larger deviations, indicating that the proposed architecture is better suited to this more nonlinear response region. LSTM generally outperforms MLP, which is consistent with the importance of temporal feature extraction.

\begin{figure}[htbp]
\centering
\includegraphics[width=\textwidth]{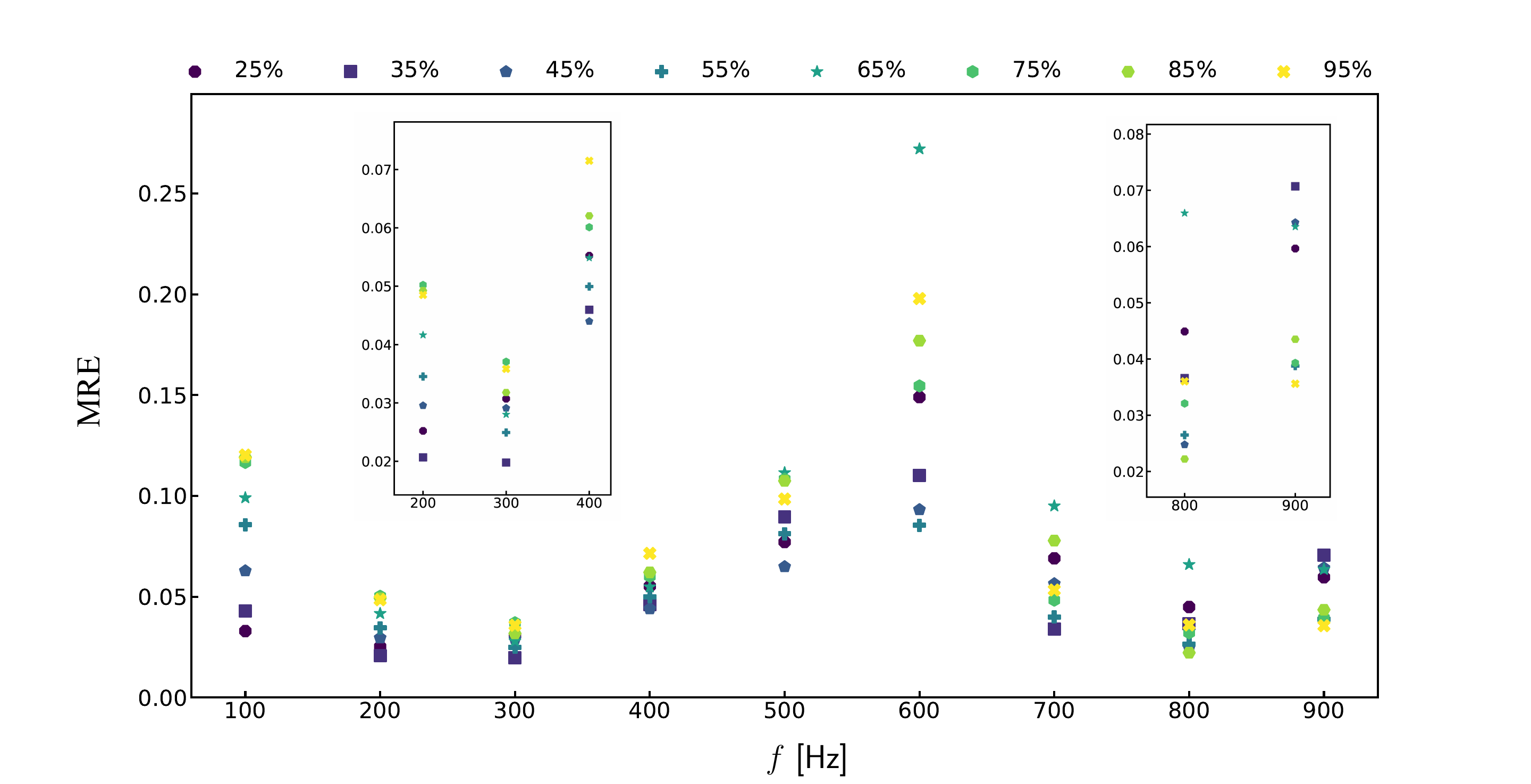}
\caption{MRE values of single-frequency signals predicted by the Dual-Path model at different amplitudes and frequencies. The value after each marker at the top of the figure denotes $A$.}
\label{result-statistics-all-amplitudes-frequencise-Dual-Path}
\end{figure}

The MRE statistics in Fig.~\ref{result-statistics-all-amplitudes-frequencise-Dual-Path} quantify prediction accuracy over a wider set of single-frequency tests. The tested amplitudes range from 0.25 to 0.95 with a step of 0.1, and each amplitude is paired with frequencies from 100 to 900~Hz with a step of 100~Hz, giving 72 signals in total. Almost all MRE values are below 10\%, and the average MRE is approximately 6.69\%. For comparison, the average MRE values of the MLP and LSTM baselines are approximately 23.85\% and 12.50\%, respectively; their detailed distributions are provided in Supplementary Figs.~S11 and S12. The computational cost also supports the proposed balance between accuracy and sequence modeling. MLP is the cheapest baseline, and the Dual-Path model requires approximately 11 times the MLP training time and 103 times the MLP inference time. Directly applying LSTM to the original velocity sequence is much more expensive, requiring approximately 205 times the MLP training time and 2727 times the MLP inference time. These results support the use of the Dual-Path model when accurate nonlinear response construction is required.
	
\begin{figure}[htbp]
\centering
\includegraphics[width=0.78\textwidth]{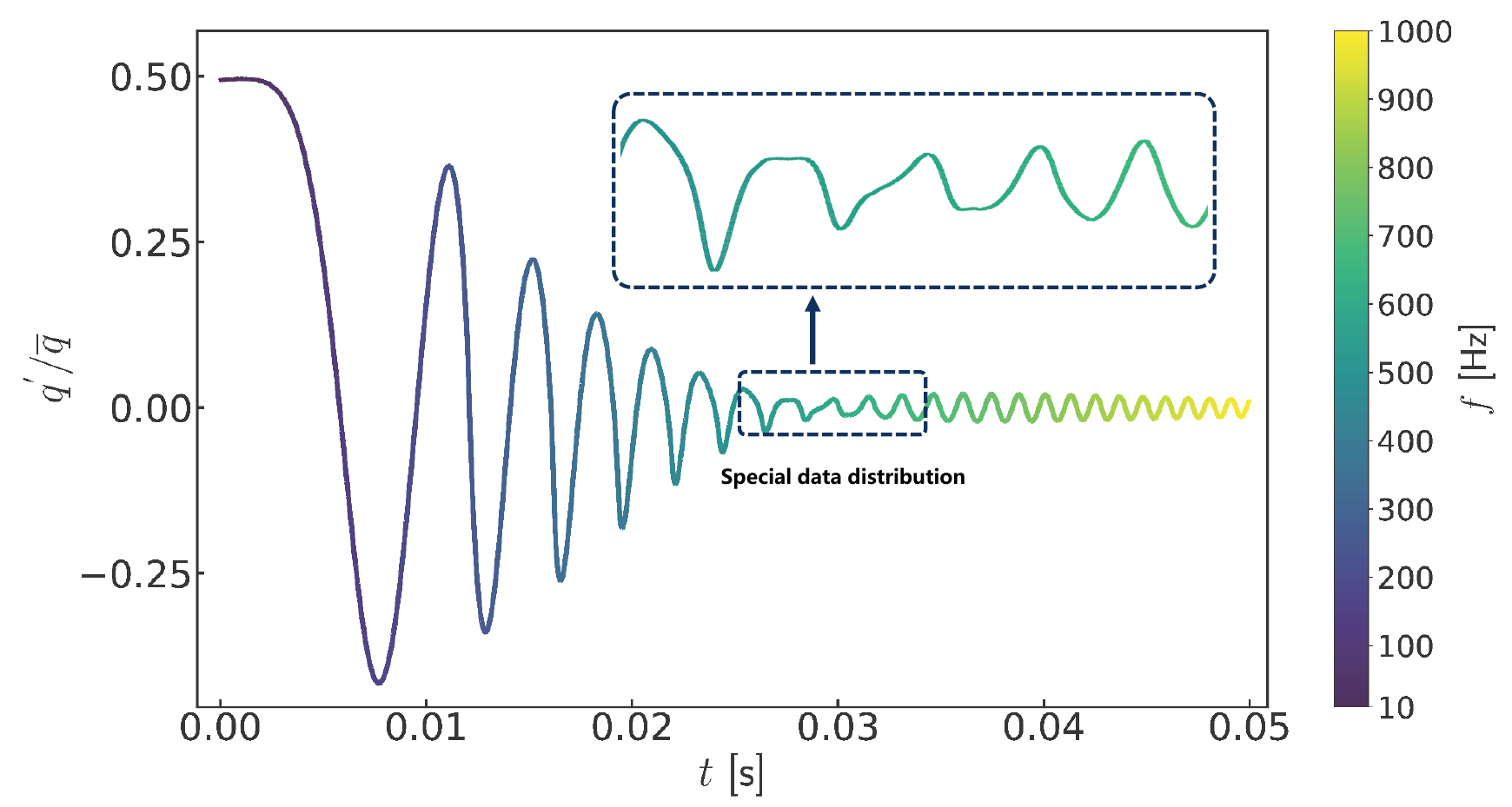}
\caption{Heat-release-rate results from the numerical simulation under a frequency-sweeping signal. The frequency-sweeping amplitude is $A=0.5$, and the frequency range is 10--1000~Hz.}
\label{special-distribution-of-frequency}
\end{figure}

Fig.~\ref{result-statistics-all-amplitudes-frequencise-Dual-Path} also shows that prediction accuracy decreases near 600~Hz. To understand this behavior, we revisited the numerical response of the 10--1000~Hz sweeping signal in Fig.~\ref{special-distribution-of-frequency}. The heat-release-rate signal exhibits a distinctive nonlinear response between 510 and 660~Hz. This regime occupies only a small fraction of the training dataset, so the network has fewer examples from which to learn this mapping. The learned response in this band can therefore be biased toward the behavior in the surrounding frequency ranges. Even so, the Dual-Path model generalizes better than MLP and LSTM in Figs.~\ref{result-0.25amplitude}, \ref{result-0.55amplitude}, and \ref{result-0.85amplitude}. Supplementary Material E compares the Dual-Path model with a Single-Path model that contains only the CFP. In the 510--660~Hz band, the Single-Path model does not achieve comparable performance, indicating that the TDFP contributes to generalization under localized data-distribution shifts. When this localized response is absent, the proposed method predicts responses accurately across the tested frequencies, as shown in Supplementary Material F.
	
\begin{figure}[htbp]
\centering
\includegraphics[width=0.9\textwidth]{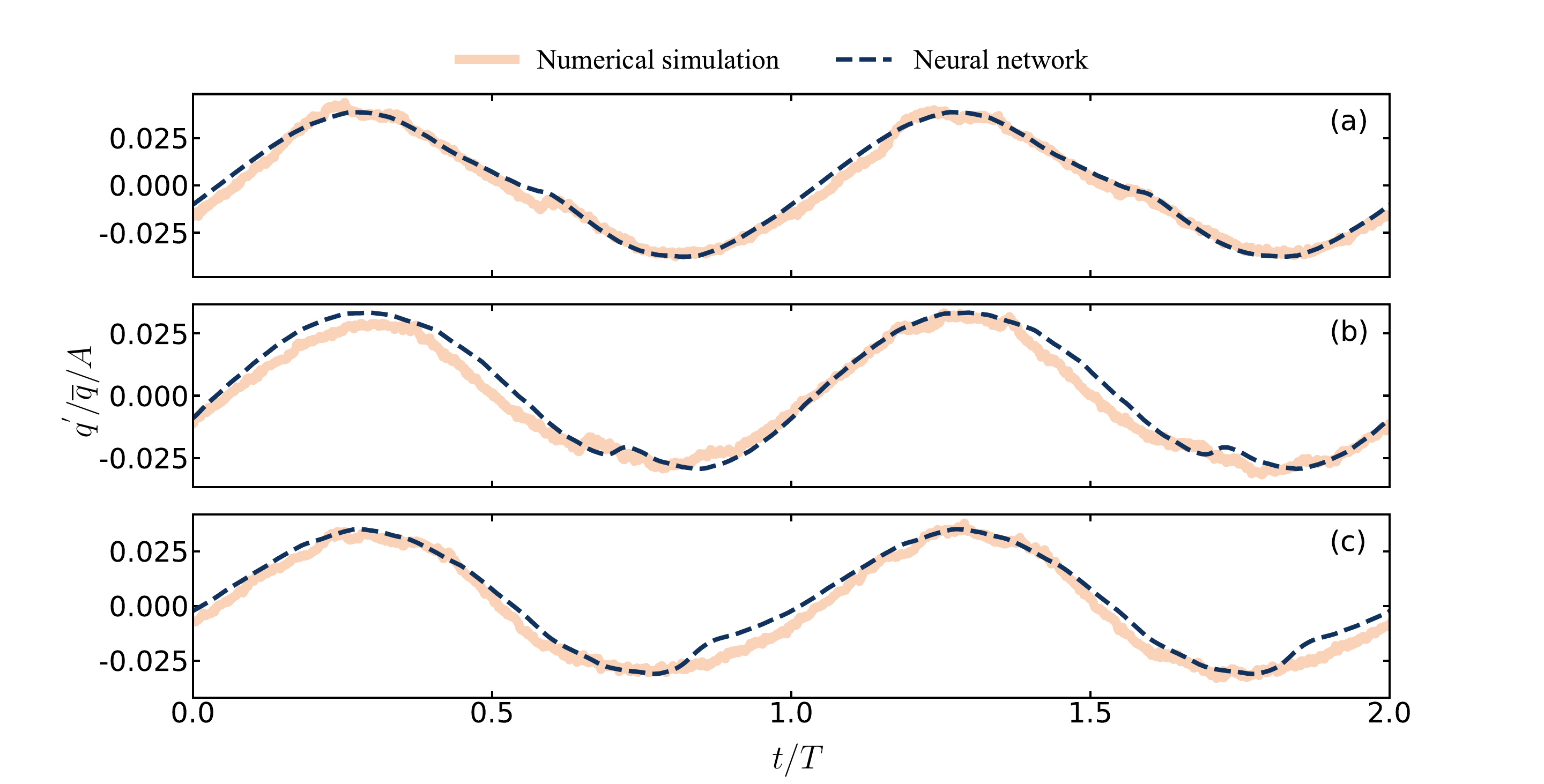}
\caption{Comparison between Dual-Path model predictions and numerical simulation results. $A=0.1$, and $f$ is 700~Hz, 800~Hz, and 900~Hz from $(a)$ to $(c)$, respectively. $n=6000$ and $n_{s}=1000$.}
\label{result-0.1amplitude-700-900Hz}
\end{figure}

\begin{figure}[htbp]
\centering
\includegraphics[width=0.78\textwidth]{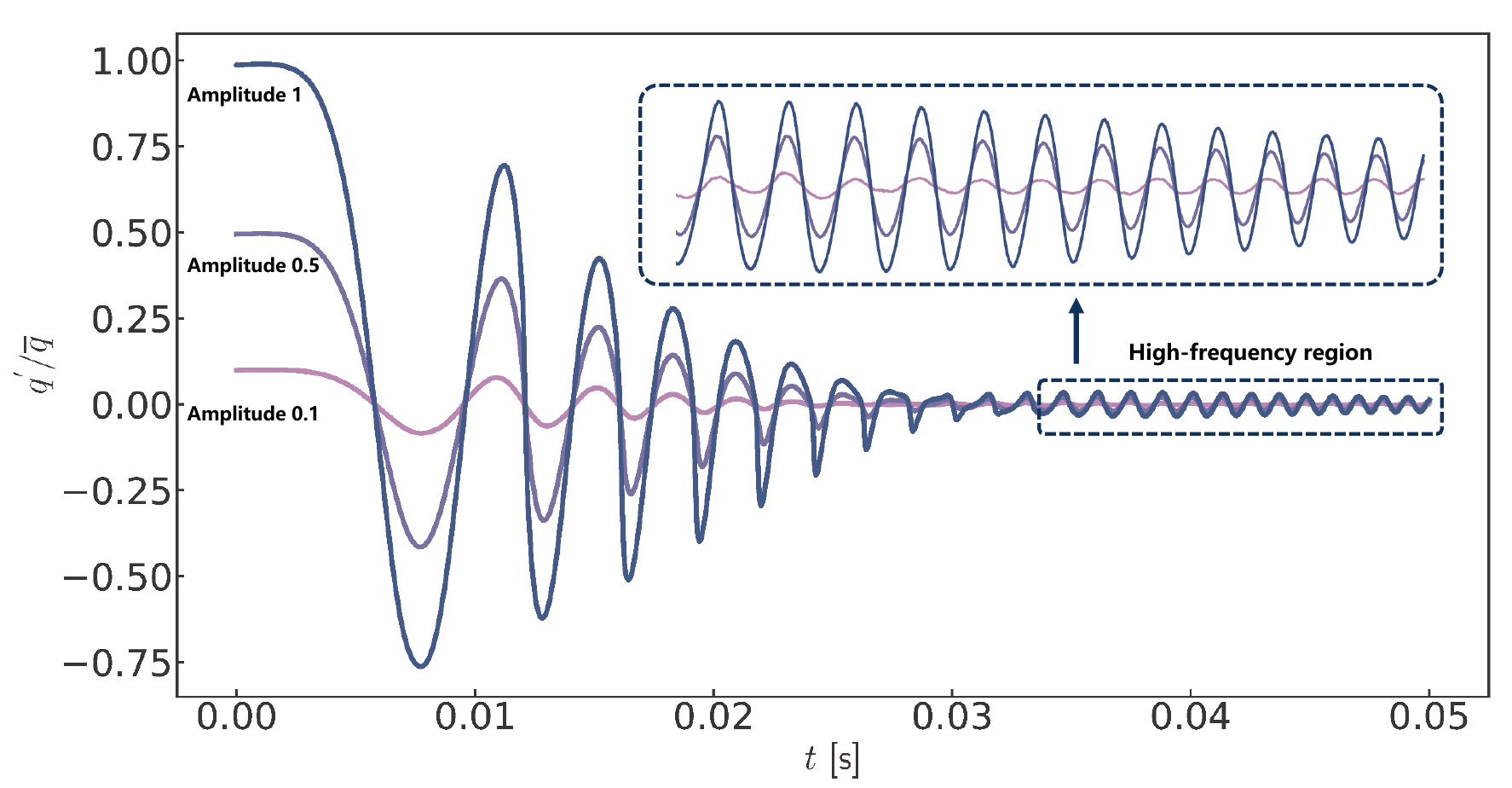}
\caption{Comparison of heat-release-rate results for frequency-sweeping signals with different amplitudes. The amplitudes are 0.1, 0.5, and 1, and the frequencies are swept from 10 to 1000~Hz.}
\label{comparison-different-heatreleaserate}
\end{figure}

A second underrepresented regime is the small-amplitude high-frequency (SAHF) response, represented here by signals with $A=0.1$ and frequencies of 700--900~Hz. Fig.~\ref{result-0.1amplitude-700-900Hz} shows that prediction accuracy deteriorates in this regime. The reason is a distribution shift in the heat-release-rate values: for $A=0.1$ at high frequency, the response amplitude is much smaller than that of the overall training data. Fig.~\ref{comparison-different-heatreleaserate} compares heat-release-rate signals for amplitudes of 0.1, 0.5, and 1 and shows the small fluctuations of the SAHF response. Even in this difficult regime, the Dual-Path model outperforms the Single-Path model, as shown in Supplementary Material E, which supports the role of the TDFP in handling specialized data distributions.
 
Transfer learning is used to improve SAHF prediction without regenerating a large training set \cite{zhuang2021comprehensive,weiss2016survey}. A model trained on one task can be adapted to a related target task using limited target-domain data. Here, the model pre-trained on the frequency-sweeping dataset in Sec.~\ref{sec:training-dataset-construction} is fine-tuned using a small amount of SAHF data.
    
\begin{figure}[htbp]
\centering
\includegraphics[width=0.82\textwidth]{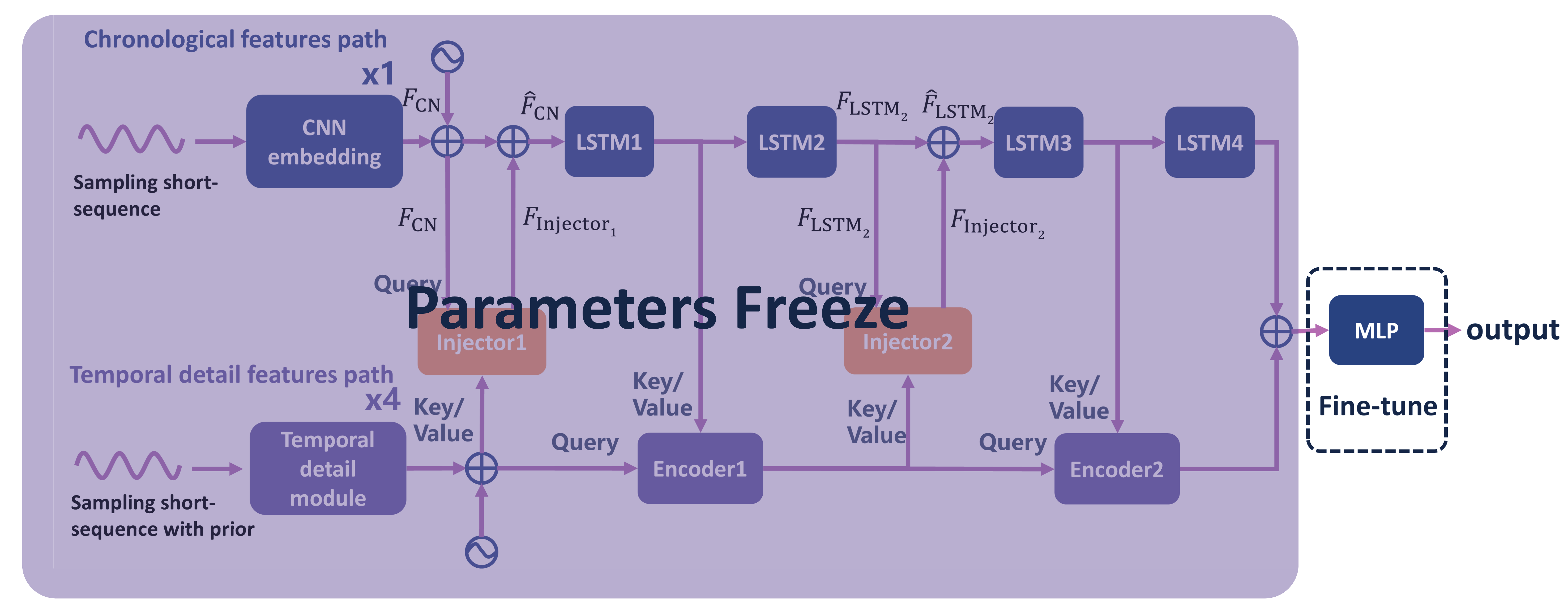}
\caption{Schematic of transfer learning.}
\label{transfer-learning}
\end{figure}

The Dual-Path model trained with 0.5 seconds of numerical simulation data has already learned the general nonlinear flame-response pattern. For SAHF adaptation, we therefore freeze all model components except the MLP module, as illustrated in Fig.~\ref{transfer-learning}. This choice preserves the broadly useful sequence features while allowing a small parameter subset to specialize for the underrepresented SAHF regime. In the example shown in Supplementary Material G, the model is fine-tuned using two 0.01~s single-frequency time series with $A=0.1$ at 800 and 900~Hz. After fine-tuning, prediction accuracy improves over the 800--900~Hz range.

This selective fine-tuning strategy uses limited target-regime data and adds little training cost. It is therefore a practical way to improve prediction accuracy when a small response regime is underrepresented in the original frequency-sweeping dataset.

\subsection{Response construction under controlled strong nonlinearity}

The preceding results assess nonlinear response construction for the numerical flame data. This subsection further tests the model under controlled strong nonlinearity. This setting differs from the localized special distribution near 510--660~Hz: here, the entire training dataset is generated from a strongly nonlinear mapping. Following our previous work~\cite{wu2023reconstruction}, the verification uses a modified classical $n-\tau$ model \cite{yang2021comparison}, which allows the level of nonlinearity to be adjusted systematically. The model is written as

    \begin{equation}
    \begin{aligned}
    \frac{q^{\prime}(t)}{\bar{q}}=\int_0^\infty h(t) y(t-t^\ast)\mathrm{d}t^\ast
    \end{aligned}
    \label{Eq: n-tau}
    \end{equation}

     where
    \begin{equation}
    \begin{aligned}
     h(t) = \mathcal{L}^{-1}(H(s)),\quad H(s) = \frac{\omega_{c}}{s+\omega_{c}},
    \end{aligned}
    \label{Eq: n-tau1}
    \end{equation}

    \begin{equation}
    \begin{aligned}
    y(t) = a_{1} \frac{u^{\prime}\left(t-\tau_{u}\right)}{\bar{u}}-a_{3}\left(\frac{u^{\prime}\left(t-\tau_{u}\right)}{\bar{u}}\right)^{3},
    \end{aligned}
    \label{Eq: n-tau2}
    \end{equation}
    and $h(t) = \mathcal{L}^{-1}(H(s))$ denotes the inverse Laplace transform of $H(s)$.

\begin{figure}[htbp]
\centering
\includegraphics[width=\columnwidth]{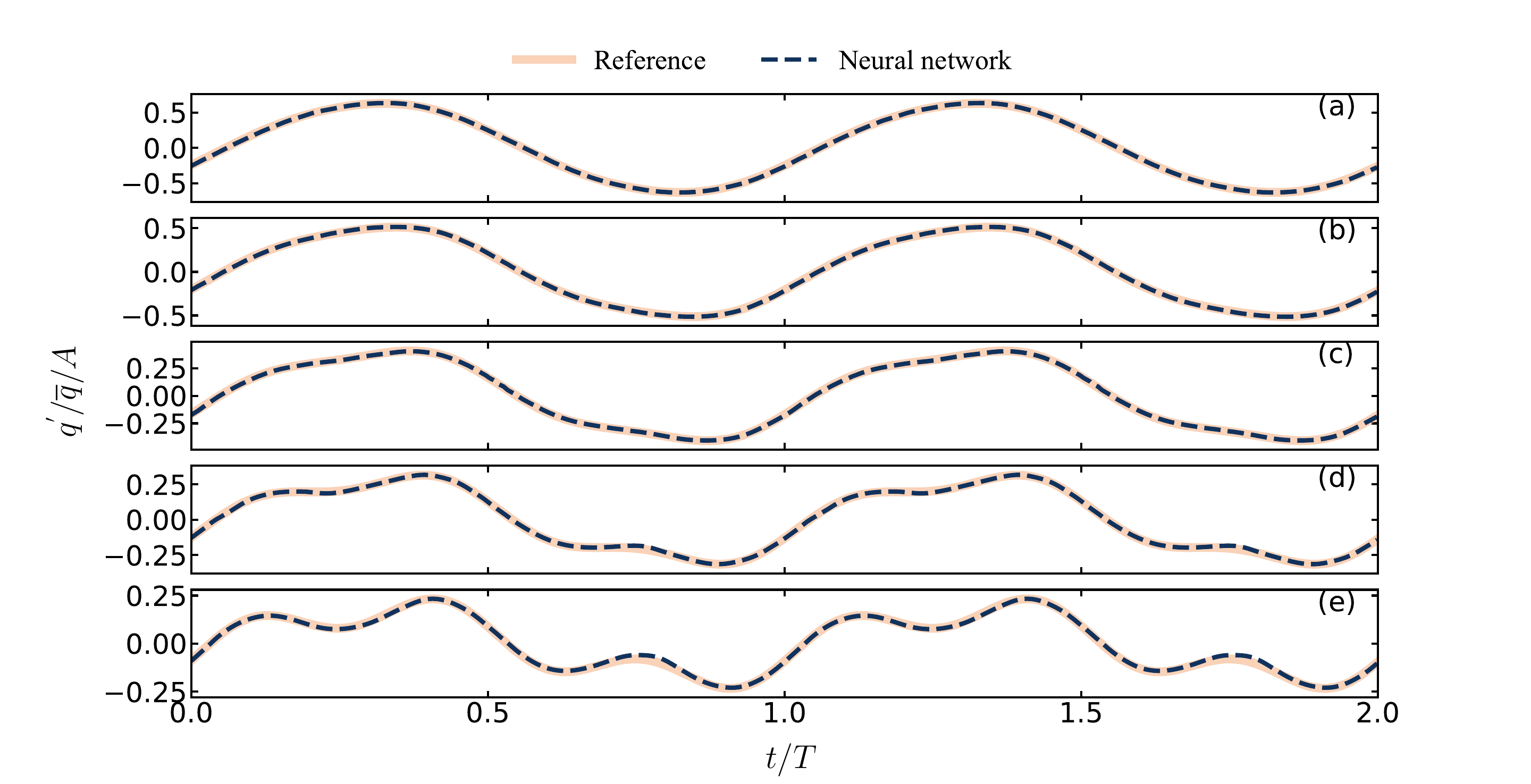}
\caption{Neural-network prediction results for different flame nonlinearity strengths. From $(a)$ to $(e)$, the ratios $a_3/a_1$ are 1, 2, 3, 4, and 5. The training dataset contains 100,000 samples. The input-signal amplitude is 0.3, and the frequency is 350~Hz. Orange solid line: results produced by Eq.~\eqref{Eq: n-tau}. Blue dashed line: results predicted by the neural network.}
\label{result-training-dataset-size-100000}
\end{figure}

The angular cutoff frequency of the first-order filter is $\omega_c = 2\pi f_c$, where $s=\mathrm{i}\omega=\mathrm{i} 2\pi f$ is the Laplace variable. The model also includes dimensionless constants $a_1$ and $a_3$ and the time delay $\tau_{u}$. In this study, $f_c=400$~Hz, $\tau_{u}=2$~ms, and $a_1 = 1$. Equation~\eqref{Eq: n-tau} is used to generate heat-release-rate signals for both training and testing. The cubic term in Eq.~\eqref{Eq: n-tau2} introduces nonlinearity. According to Noiray et al., this form represents nonlinear behavior when $\frac{u^{\prime}\left(t-\tau_{u}\right)}{\bar{u}}<\sqrt{\frac{a_{1}}{2a_{3}} }$ \cite{noiray2011investigation}. The nonlinearity strength is adjusted through the ratio $a_3/a_1$. We set $a_3/a_1$ to 1, 2, 3, 4, and 5, with the maximum capped at 5 to prevent the cubic term from overwhelming the physical interpretation of the model. Fig.~\ref{result-training-dataset-size-100000} shows the prediction results for these nonlinearity levels, with each level trained independently.

\begin{figure}[htbp]
\centering
\includegraphics[width=\columnwidth]{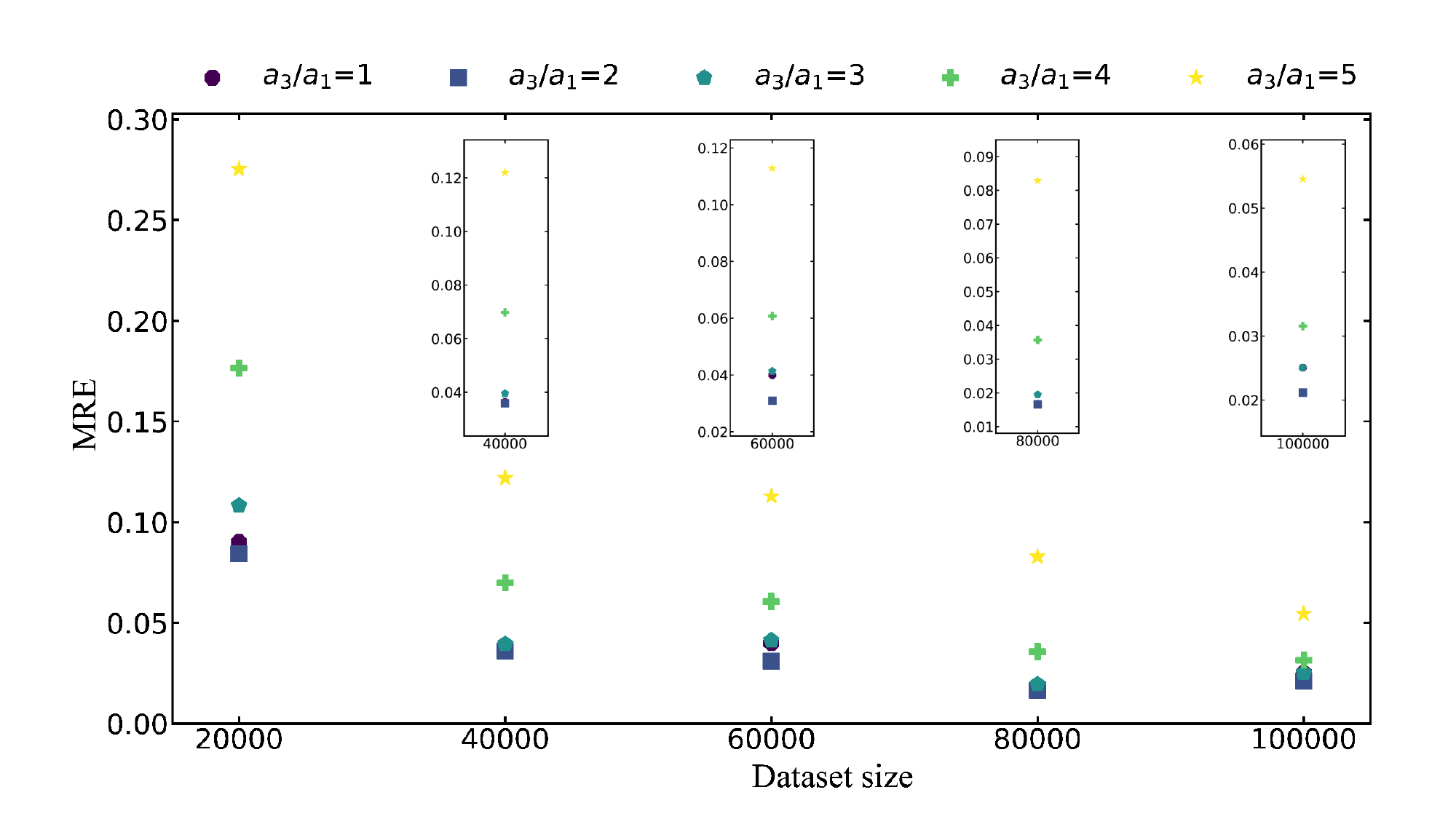}
\caption{MRE statistics for different nonlinearity strengths and training dataset sizes. The nonlinearity strengths are 1, 2, 3, 4, and 5. The training dataset sizes are 20,000, 40,000, 60,000, 80,000, and 100,000.}
\label{result-statistics-sizes-intensities}
\end{figure}

The model reconstructs the response over the tested nonlinearity range, but the data requirement depends on nonlinearity strength. Fig.~\ref{result-statistics-sizes-intensities} shows the average MRE for 10 single-frequency signals with frequencies from 100 to 1000~Hz and an amplitude of 0.3. The horizontal axis is the training dataset size, and the vertical axis is the average MRE between the neural-network prediction and the reference model. At a fixed dataset size, the MRE increases with nonlinearity strength. At a fixed nonlinearity strength, the MRE decreases as the dataset size increases. Thus, stronger nonlinearities require more data to resolve the underlying mapping, whereas weaker nonlinearities can be modeled with smaller datasets.

\section{Conclusion}

This paper develops a data-driven approach for efficiently constructing nonlinear flame thermoacoustic response in the time domain. The motivation is the need for compact flame-response models that can support nonlinear combustion-instability analysis in propulsion combustor design without requiring a separate high-fidelity simulation for every forcing frequency and amplitude. The proposed framework combines a frequency-sweeping training dataset, short-sequence sampling, and a Dual-Path surrogate architecture that extracts both ordered sequence features and temporal-detail features.

The numerical validation shows that the proposed training strategy and surrogate architecture can reconstruct nonlinear flame response from limited simulation data. For 72 single-frequency test signals covering amplitudes from 0.25 to 0.95 and frequencies from 100 to 900~Hz, the Dual-Path model achieves an average MRE of approximately 6.69\%, compared with 23.85\% for MLP and 12.50\% for LSTM baselines. The short-sequence sampling strategy reduces the original sequence length from 6000 to 1000 and increases training and inference speeds by factors of 15.4 and 7.7, respectively. Additional tests with the modified $n-\tau$ model show that stronger nonlinearities require larger training datasets, providing a practical guideline for balancing simulation cost and response-construction accuracy.

The principal implication of this study is methodological rather than engine-specific. The proposed model can be viewed as a step toward a data-driven flame-response closure for low-order thermoacoustic analysis of propulsion combustors. It is most useful when nonlinear response construction is required; if only linear flame response is needed, classical system-identification-based FTF determination remains more economical and physically direct.

Several limitations define the scope of the present study. The training and validation data are generated from a single laminar premixed flame at one operating condition, so robustness under turbulence-induced fluctuations, measurement noise, high pressure, multiphase injection, and engine-scale geometry has not yet been established. The strong-nonlinearity test uses a cubic $n-\tau$ model and therefore does not represent all possible higher-order nonlinear flame dynamics. Model accuracy also depends on how well the training data cover the target response regime; underrepresented regions, such as small-amplitude high-frequency responses, can reduce prediction accuracy. Future work will extend the framework to noisy and turbulent datasets, multiple flame configurations, and propulsion-relevant operating conditions, and will couple the learned flame response with acoustic network models for nonlinear stability prediction.

\section*{Supplementary material}
See the supplementary material for neural-network architectures and configurations, additional comparisons supporting the chronological-feature and generalization analyses, and supplementary results on the transfer-learning capability of the Dual-Path model.

\section*{CRediT authorship contribution statement}
Jiawei Wu: Conceptualization (equal), Investigation (equal), Methodology (equal), Writing - original draft (equal).
Teng Wang: Data curation (lead).
Jiaqi Nan: Writing - review \& editing (equal).
Wang Han: Conceptualization (equal).
Jingxuan Li: Conceptualization (equal), Funding acquisition (equal), Writing - review \& editing (equal).
Lijun Yang: Funding acquisition (equal).

\section*{Data availability}
The code supporting this study is available in the \href{https://github.com/wujiaweii/Constructing-the-flame-nonlinear-response/tree/main}{GitHub repository}. The data supporting this study are available in the \href{https://zenodo.org/records/15276195}{Zenodo record}.

\section*{Declaration of competing interest}
The authors declare that they have no known competing financial interests or personal relationships that could have appeared to influence the work reported in this paper.

\section*{Acknowledgments}
The authors gratefully acknowledge financial support from the National Natural Science Foundation of China (Grant No. 52376089) and Beijing Natural Science Foundation (Grant No. L241003).

\bibliographystyle{elsarticle-num}
\bibliography{reference}
\end{document}